\documentclass[10pt,twocolumn,letterpaper]{article}

\usepackage{multirow}
\usepackage[pagenumbers]{cvpr} 
\definecolor{cvprblue}{rgb}{0.21,0.49,0.74}
\usepackage[pagebackref,breaklinks,colorlinks,allcolors=cvprblue]{hyperref}

\def\confName{CVPR}
\def\confYear{2027}

\title{Multi-Granularity Context-Enhanced RAG over Multimodal Knowledge Graphs}

\author{
    Zongyu Wu$^{1*}$ \quad
    Yilong Wang$^{1*}$ \quad
    Xiaochen Wang$^1$ \quad
    Minhua Lin$^1$ \quad
    Zhichao Xu$^2$ \\
    Fenglong Ma$^1$ \quad
    Xiang Zhang$^1$ \quad
    Suhang Wang$^1$ \\
    $^1$The Pennsylvania State University \qquad
    $^2$Independent Researcher \\
    {\tt\small
    \{zongyuwu,szw494\}@psu.edu}
}

\begin{document}
\maketitle

\begingroup
\renewcommand{\thefootnote}{\fnsymbol{footnote}}
\footnotetext[1]{Core contributors.}
\endgroup

\begin{abstract}
Knowledge graph (KG)-based retrieval-augmented generation (RAG) leverages structured knowledge to provide large language models (LLMs) with high-quality external information. Building on these works, recent studies have explored multimodal knowledge graphs (MMKGs) as knowledge bases for GraphRAG. This enables GraphRAG to integrate knowledge across multiple modalities, thereby further enhancing its performance. However, existing MMKG-based RAG methods generally follow a common pipeline in which different modalities are largely processed independently before being fused. As a result, textual context is only used to a limited extent during visual information extraction and subsequent multimodal knowledge fusion. This brings a semantic gap between images and text which limits the multimodal GraphRAG performance. To address this issue, we propose a novel framework for constructing a Context-Enhanced MMKG (CEMMKG) to better support multimodal GraphRAG. The proposed CEMMKG enriches each image with complementary textual context at both local and global scopes. Local context goes beyond the surrounding text by incorporating sentences that are semantically related to the image, while global context provides a summary of the entire document. We further introduce a multi-granularity design for the local context, allowing it to capture semantically relevant information at different levels of detail. Extensive experiments on multiple datasets validate that CEMMKG is effective in leveraging contextual information to improve MMKG-based RAG performance. Moreover, its effectiveness across different MMKG-based RAG methods demonstrates its broad applicability.
\end{abstract}    
\section{Introduction}
\label{sec:intro}

\begin{figure}[htbp]
  \centering
 \includegraphics[width=1\linewidth]{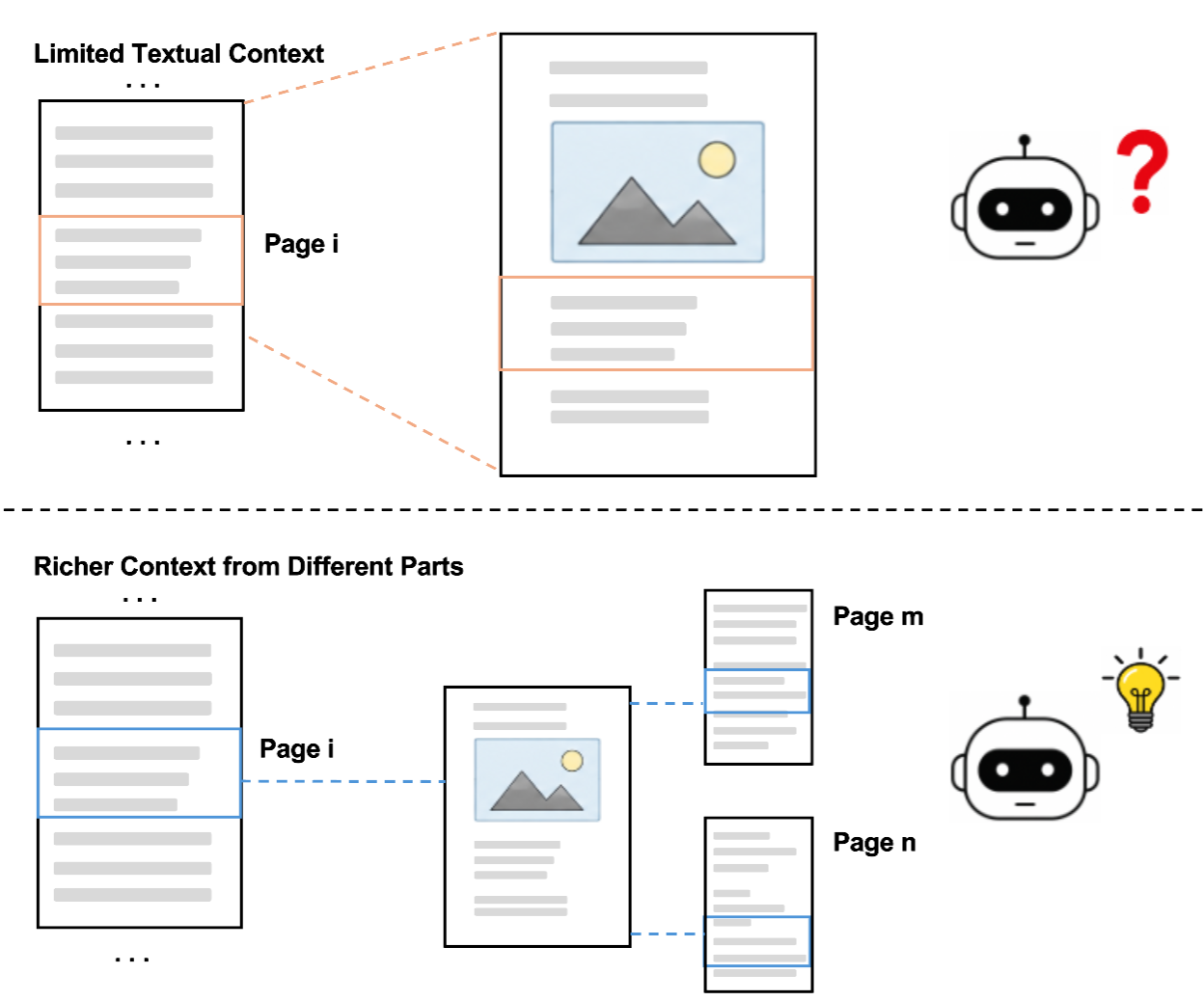}
  \caption{An illustration of textual context for a selected visual element, where useful contextual information may be distributed across different parts of a document rather than being limited to the immediate surroundings of the visual element.}
  \label{fig:contextmotivation}
\end{figure}

Although Multimodal Large Language Models (MLLMs)~\citep{liu2023llava,liu2024llava15,yin2024mllmsurvey} have demonstrated strong performance across a wide range of domains, they are still prone to generate hallucinated content~\citep{bai2024mllmhallusurvey} due to several factors such as outdated training knowledge. Retrieval-Augmented Generation (RAG), originally developed for LLMs~\citep{Gao2023RAGSurvey,Lewis2020RAG,xu2026VERITAS}, has been extended to MLLMs~\citep{abootorabi2025mmragsurvey}. By retrieving relevant information from external data sources~\citep{abootorabi2025mmragsurvey}, RAG can augment MLLMs with external knowledge which can support response generation. Among RAG approaches, GraphRAG~\citep{peng2025graphragsurvey} has gained increasing attention for leveraging structured data, such as knowledge graphs (KGs)~\citep{ji2022kgsurvey}, as external knowledge sources. By explicitly modeling entities and the relations among them, KGs provide structured knowledge that can effectively support information retrieval and model generation.

Despite the success of KGs to represent knowledge, most existing GraphRAG methods~\citep{edge2024graphrag} primarily operate on textual KGs, with graph construction and retrieval largely centered on textual information. This text-centric paradigm leaves rich information from other modalities, such as visual content and tabular data, largely underexplored. Information from different modalities could provide valuable knowledge beyond text, enabling a more comprehensive representation of KG. To incorporate multimodal information into KGs, recent works such as RAG-Anything~\citep{guo2025raganything} and MMGraphRAG~\citep{wan2025mmgraphrag} explore the construction of retrieval-oriented multimodal knowledge graphs (MMKGs) and modality fusion strategies for multimodal GraphRAG. An MMKG represents information from different modalities as nodes within a unified graph.

However, existing MMKG-based RAG methods~\citep{wan2025mmgraphrag,guo2025raganything} typically follow a pipeline in which information from each modality is largely processed separately before being fused into a unified MMKG. Such modality-specific processing may overlook the rich contextual relationships between information from different modalities. For example, when processing visual information, existing methods often utilize only limited textual context, such as surrounding text chunks. However, visual information is often closely related to textual information distributed across different parts of a document as shown in~\cref{fig:contextmotivation}. Such textual information may provide important context for interpreting the semantics of visual elements. Limited textual context may result in suboptimal visual knowledge extraction and ineffective cross-modal knowledge fusion, thereby constraining the performance of MMKG-based RAG. 

Addressing this limitation requires more than simply incorporating additional textual context. Different context granularities capture information at varying levels of detail, and their effectiveness may depend on how the context is utilized during the MMKG construction process. These considerations raise three important yet underexplored questions: (\textbf{i}) what textual information should be selected as context for visual elements, (\textbf{ii}) what level of granularity is most appropriate for constructing such context, and (\textbf{iii}) how the designed context can be effectively utilized during MMKG construction.

To answer these questions, we propose CEMMKG, a multi-granularity context-enhanced MMKG construction framework that systematically investigates the design and utilization of textual context for visual elements. Specifically, we first investigate different context designs and granularities, considering multiple granularities of local context alongside global context to facilitate MMKG construction. We then explore how the resulting context can be effectively leveraged at different stages of MMKG construction. Extensive experiments across multiple datasets and context configurations demonstrate the effectiveness of CEMMKG in improving MMKG-based RAG performance. Furthermore, the proposed CEMMKG can be effectively integrated with different MMKG-based RAG methods, demonstrating its broad applicability.
\section{Related Work}
\label{sec:related_work}

\subsection{Multimodal RAG}
Multimodal RAG~\citep{chen2022murag,yasunaga2023racm3,suri2025visdom,yeo2026universalrag} extends retrieval-augmented generation beyond text, so that the information in other modalities such as images and charts can also ground generation~\citep{abootorabi2025mmragsurvey}. Existing methods improve multimodal RAG through different designs such as late interaction over fine-grained visual representations~\citep{lin2023flmr}, hierarchical retrieval~\citep{caffagni2024wikillava}, and evidence filtering~\citep{ling2025mmkbrag}. For visually rich documents, approaches either retrieve rendered document pages directly~\citep{yu2025visrag,faysse2025colpali,cho2024m3docrag} or convert documents into text for conventional retrieval~\citep{wang2024mineru}. While effective, these approaches primarily focus on retrieving evidence rather than explicitly modeling relationships among multimodal information. Moreover, the model may fail to fully leverage the retrieved evidence when it is simply concatenated~\citep{liu2026primacybias}.

\subsection{GraphRAG} 
GraphRAG addresses the fragmentation and redundancy in conventional RAG by retrieving over structured knowledge~\citep{peng2025graphragsurvey,han2024graphragsurvey2,wang2024kgp,edge2024graphrag,guo2025lightrag,gutierrez2024hipporag,sarthi2024raptor,chen2026pathrag,liang2025kag,wang2026gpr} such as community-based retrieval~\citep{edge2024graphrag}. Recent multimodal GraphRAG methods extend this paradigm by constructing multimodal knowledge graphs (MMKGs) to support structure-aware multimodal retrieval~\citep{wan2025mmgraphrag,guo2025raganything,yuan2026mkgrag,hsiao2026megarag,park2026m3kgrag,dai2026mg2rag,he2026hvmgraphrag,wang2026mkgragbench}. While these methods demonstrate the potential of MMKG-based retrieval, their effectiveness
heavily depends on the quality of the underlying MMKG,
particularly how visual and textual information are
connected during MMKG construction.

\subsection{Multimodal Knowledge Graph Construction}
MMKGs extend traditional knowledge graphs~\citep{ji2022kgsurvey}
by incorporating multimodal information~\citep{
zhu2024mmkgsurvey,chen2024kgmmsurvey,park2025vatkg}.
Existing MMKGs can be broadly categorized into two types
based on representations. \textit{Attribute-based} MMKGs attach images to symbolic entities as an additional attribute~\citep{liu2019mmkg,wang2020richpedia,alberts2021visualsem}, without explicitly modeling the fine-grained content within each image, whereas \textit{node-based} MMKGs promote visual content to first-class nodes so that objects and their relations are explicitly modeled and traversable~\citep{dai2026mg2rag,wang2026mkgragbench}. Recent multimodal GraphRAG methods~\citep{wan2025mmgraphrag,guo2025raganything} mainly focus on node-based MMKGs constructed from multimodal documents, where textual context is important because visual elements are often difficult to fully understand in isolation.

However, existing approaches~\citep{wan2025mmgraphrag,guo2025raganything} often construct visual and textual graph structures largely separately and integrate them at a later stage. During visual information processing,
some approaches process visual content without accompanying textual context, while others rely on a fixed window of positionally adjacent text chunks. Such limited or proximity-based context utilization may introduce irrelevant information while overlooking semantically relevant texts elsewhere in the document, potentially compromising cross-modal grounding and MMKG quality. Despite the importance of textual context, how different forms of context contribute to MMKG construction remains insufficiently understood.

To address these limitations, we first systematically define textual context at different levels of granularity for visual information processing and cross-modal fusion. We then comprehensively investigate how different forms of textual context can be leveraged to support MMKG construction.
\section{Background and Preliminaries}
In this section, we start from the general formulation of multimodal knowledge graph (MMKG)-based RAG and then progressively focus on the specific problem studied in this work.

\subsection{MMKG-based RAG}
\label{sec:formulation}
MMKG-based RAG~\citep{wan2025mmgraphrag} extends conventional GraphRAG to support knowledge sources spanning multiple modalities, such as text and images. Given a multimodal document $\mathcal{D}$, its textual content $\mathcal{T}$ and visual elements $\mathcal{I}={I_1,\ldots,I_N}$ can be extracted using a document parser such as MinerU~\citep{wang2024mineru}. An MMKG-RAG system then constructs an MMKG $\mathcal{G}_M=(\mathcal{V},\mathcal{E})$ based on $\mathcal{D}$, where $\mathcal{V}$ and $\mathcal{E}$ denote the sets of entities and relations, respectively. Given a user query $q$, the relevant multimodal knowledge in $\mathcal{G_{M}}$ is then retrieved and provided to an LLM or MLLM $\mathcal{F}_{\mathrm{model}}$ as external evidence to produce a more accurate response:
\begin{equation}
    \hat{y} = \mathcal{F}_{\mathrm{model}}\big(q, \mathcal{R}(q,\mathcal{G_{M}})\big),
\label{eq:mmkg_rag}
\end{equation}
where $\mathcal{R}$ denotes the retriever and $\hat{y}$ is the generated answer.

\subsection{Multimodal Knowledge Graph Construction}
\label{sec:construction}
Eq.~\eqref{eq:mmkg_rag} highlights two key factors that influence the quality of the final answer: the retriever $\mathcal{R}$ and the underlying knowledge graph $\mathcal{G}_M$ on which it operates. Since retrieval can only operate on the knowledge represented in the graph, the quality of $\mathcal{G}_M$ plays a fundamental role in the overall performance of MMKG-based RAG. Accordingly, we focus on the construction of $\mathcal{G}_M$ from the document $\mathcal{D}$, rather than on the retrieval stage.

Existing MMKG-RAG methods construct $\mathcal{G_{M}}$ by processing each modality almost separately and merging the results: a text-based graph $\mathcal{G}^{\mathrm{t}}$ is built from $\mathcal{T}$ and an image-based graph $\mathcal{G}^{\mathrm{v}}$ from $\mathcal{I}$, independently of each other. The two graphs are then integrated into a unified MMKG:
\begin{equation}
\mathcal{G_{M}}
=
f_{\mathrm{fusion}}
\left(
\mathcal{G}^{\mathrm{t}},
\mathcal{G}^{\mathrm{v}}
\right),
\label{eq:generic_fusion}
\end{equation}
where $f_{\mathrm{fusion}}$ is a modality fusion module which can identify entities of $\mathcal{G}^{\mathrm{v}}$ and $\mathcal{G}^{\mathrm{t}}$ that denote the same underlying object and then pair them. For $f_{\mathrm{fusion}}$, we follow the fusion procedure proposed in previous work~\citep{wan2025mmgraphrag} and keep its overall workflow fixed throughout this study. For each visual entity, candidate textual entities are collected from neighboring text chunks and partitioned by clustering, after which an LLM identifies its textual counterpart from the most relevant cluster. Unmatched visual entities are enriched with textual context, and the resulting cross-modal alignments are used to fuse the image and text KGs into a unified MMKG. Constructing $\mathcal{G}_M$ from $\mathcal{D}$ therefore involves constructing the textual and visual branches and subsequently fusing them.

Textual graph construction inherits the well-established pipeline of text-based GraphRAG, whereas vision-based graph construction encounters a greater challenge, as visual information is more susceptible to information loss during its transformation into graph representations. Information not captured during this process may be unavailable to subsequent stages, including modality fusion and information retrieval. We therefore further narrow our focus to the construction of the vision-based graph $\mathcal{G}^{\mathrm{v}}$, while also considering its subsequent integration with the textual graph through modality fusion.

Existing methods consider such textual context only to a limited extent when building $\mathcal{G}^{\mathrm{v}}$, primarily using text chunks located near the image in the image-to-graph module. However, richer contextual information that is semantically relevant to the image may be distributed across other parts of the document and remains largely underexplored.

The problem studied in this work can be stated as follows: \textit{given a multimodal document $\mathcal{D}$ with textual content $\mathcal{T}$ and a set of visual elements $\mathcal{I}$, how can we establish meaningful and comprehensive textual context from $\mathcal{T}$ for each visual element $I_i \in \mathcal{I}$ and effectively utilize such context to construct a higher-quality MMKG?}

\section{Method}

\begin{figure*}[t]
  \centering
 \includegraphics[width=1\linewidth]{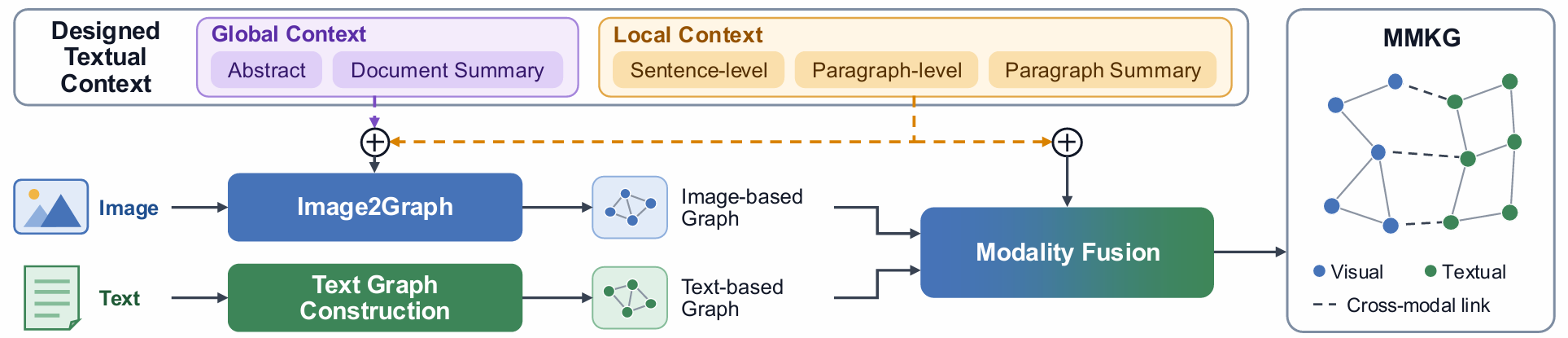}
  \caption{An overview of the designed textual context in our work. Given a target image and texts in the document, we design different levels of contextual information, including local context with different granularities and global context. The designed textual contexts can be incorporated into both the Image2Graph and modality fusion stages, individually or jointly. For clarity, only one image-based graph is illustrated in the figure.}
  \label{fig:framework}
\end{figure*}

In this section, we present CEMMKG, our proposed framework for the problem defined in~\cref{sec:construction}. Given a visual element of a document, CEMMKG constructs a textual context for it and then utilizes that context across the stages of MMKG construction. An overview of the framework is shown in~\cref{fig:framework}. ~\cref{subsec:context} addresses what textual information should be selected as context for a visual element and at what granularity it should be organized, and ~\cref{subsec:utilization} addresses how the resulting context is utilized during MMKG construction. Throughout this section, we reuse the textual content $\mathcal{T}$ and visual content set $\mathcal{I}$ introduced in ~\cref{sec:formulation}, and additionally leverage their internal structure recovered by the parser. The textual content $\mathcal{T}$ is an ordered sequence of paragraphs, with each paragraph being an ordered sequence of sentences, and we use $\mathcal{S}$ to denote the set of all sentences in $\mathcal{T}$. Each visual element $I_i \in \mathcal{I}$ carries an identifier $\mathrm{id}(I_i)$ recovered from its caption, such as \textit{Figure 3} or \textit{Table 2}.

\subsection{Multi-Granularity Textual Context}
\label{subsec:context}
We construct the context from two complementary types of contextual information, distinguished by which part of $\mathcal{T}$ they draw upon: \textbf{\textit{local context}}, which captures fine-grained details tied to a specific visual element, and \textbf{\textit{global context}}, which captures the overall content of the document at a coarse-grained level. Next, we will introduce more details about each type.

\subsubsection{Local Context}
For each visual element $I_i$, we define its local context as textual information drawn from a specific region of $\mathcal{T}$, such as a set of sentences within a certain paragraph. Local context focuses on specific details of the document rather than its overall content, and is particularly important for understanding visual information, since figures and tables often cannot be fully interpreted in isolation. We draw it from two complementary sources: the text that immediately \emph{surrounds} a visual element, and the text that explicitly \emph{refers} to it.

The latter source requires locating the sentences that mention the target element. A visual element is typically introduced at one point of a document and then discussed in detail elsewhere, and the discussing text is often far more informative about it than its positional neighbors. Using the identifier $\mathrm{id}(I_i)$ recovered during document parsing, we therefore define the \emph{reference set} of $I_i$ as
\begin{equation}
    \mathcal{R}_i
    = \left\{\, s \in \mathcal{S}
      \;\middle|\; \mathrm{id}(I_i) \text{ is mentioned in } s \,\right\}.
\label{eq:reference_set}
\end{equation}
That is, all sentences that contain a textual reference to $I_i$, such as ``\textit{Figure 3 shows $\ldots$}''. Note that $\mathcal{R}_i$ is determined by mention rather than by position. Hence, its elements may lie far away from $I_i$ in the document.

On this basis, we define the components of local context. The first is inherited from previous work and always retained; the remaining three are reference-based and describe the same reference set at different scopes and information densities:
\begin{itemize}[leftmargin=*]
    \item \textit{Surrounding Text} $c^{L}_{i,\mathrm{surr}}$: the text chunks lying immediately before and after $I_i$ in the document. Previous work~\citep{wan2025mmgraphrag,guo2025raganything} has demonstrated the importance of involving the text surrounding an image as meaningful context of the image, since a visual element and its adjacent narrative are usually introduced together. To this point, we involve it as part of the local context to provide more comprehensive information concerning the image. This component captures the local context used by existing methods, as discussed in ~\cref{sec:construction}, while our overall context design further incorporates additional richer contextual information beyond this local scope.
    \item \textit{Reference Sentence} $c^{L}_{i,\mathrm{sent}}$: for every sentence in $\mathcal{R}_i$, that sentence together with the sentences immediately preceding and following it. This is the most concise form of reference-based context, providing fine-grained textual information that directly describes or complements the information presented in the corresponding image.
    \item \textit{Reference Paragraph} $c^{L}_{i,\mathrm{para}}$: the full paragraphs in which the sentences of $\mathcal{R}_i$ appear. Paragraphs cover a broader scope, offering additional background and supporting information that may not be fully captured by the reference sentences alone, at the cost of a lower information density.
    \item \textit{Reference Paragraph Summary} $c^{L}_{i,\mathrm{sum}}$: a summary of the reference paragraphs generated by a large language model (LLM), which is designed to retain the information relevant to the target image while filtering out redundant content from the original paragraphs. It thus sits between the two forms above, seeking the coverage of a paragraph at a density closer to that of a sentence.
\end{itemize}

Since the three reference-based components provide contextual information for the same references at different granularities, they are considered alternative context configurations rather than being used jointly. Let $c^{L}_{i,\mathrm{ref}}$ denote the selected reference-based component. The local context of $I_i$ is then defined as:
\begin{equation}
    c^{L}_i = \left\{ c^{L}_{i,\mathrm{surr}},\; c^{L}_{i,\mathrm{ref}} \right\},
    \quad
    c^{L}_{i,\mathrm{ref}} \in \left\{ c^{L}_{i,\mathrm{sent}},\; c^{L}_{i,\mathrm{para}},\; c^{L}_{i,\mathrm{sum}} \right\}.
\label{eq:local_context}
\end{equation}

Because a visual element may be referenced multiple times throughout a long document, $\mathcal{R}_i$ can become large, resulting in a correspondingly long context. As overly long context degrades how effectively (M)LLMs utilize the supplied evidence~\citep{liu2024lostinthemiddle}, we bound the local context in two ways. First, we retain only the leading reference sites in document order, and use fewer of them at the fusion stage than at the image-to-graph stage, since alignment needs only enough text to name the entities involved. Second, we cap the length of the reference-based component and that of the global context separately, so that neither can crowd out the other. 

\subsubsection{Global Context}
We define global context as textual information derived from the entire textual content $\mathcal{T}$. In contrast to the previously defined local context, global context refers to a document-level semantic representation that captures the overarching knowledge of the entire document. Global context could provide broader semantic information that might help interpret visual elements beyond local-level information and facilitates cross-modal knowledge integration. To be specific, we define the following two alternative forms of global context:

\begin{itemize}[leftmargin=*]
    \item \textit{Abstract} $c^{G}_{\mathrm{abs}}$: the document's original abstract extracted from $\mathcal{T}$ and directly used as global context.
    \item \textit{Document Summary} $c^{G}_{\mathrm{sum}}$: a summary of the whole textual content $\mathcal{T}$ generated by an LLM, which captures and integrates information distributed throughout the document into a unified summary.
\end{itemize}

Unlike the local components, which are used jointly, the two global forms are interchangeable and are determined by document availability: we use $c^{G}_{\mathrm{abs}}$ for documents that provide an abstract and fall back to $c^{G}_{\mathrm{sum}}$ otherwise. Since many real-world documents fall outside the academic domain and may not contain an abstract, the latter case is common in practice. We use $c^{G}$ to denote the selected form of global context. Note that $c^{G}$ is defined at the document level and is therefore shared by all visual elements in $\mathcal{D}$, whereas $c^{L}_i$ is specific to $I_i$.

\subsubsection{Constructed Context}
Combining the two components above, the context constructed for visual element $I_i$ is
\begin{equation}
    \widetilde{C}_i
    = f_{\mathrm{context}}(I_i, \mathcal{T})
    = \left\{ c^{L}_i,\; c^{G} \right\},
\label{eq:context_def}
\end{equation}
where $f_{\mathrm{context}}$ denotes the context construction procedure defined in this subsection. Every component of $\widetilde{C}_i$ is grounded in the textual content of the document: $c^{L}_{i,\mathrm{surr}}$, $c^{L}_{i,\mathrm{sent}}$, $c^{L}_{i,\mathrm{para}}$ and $c^{G}_{\mathrm{abs}}$ are subsequences of $\mathcal{T}$, whereas $c^{L}_{i,\mathrm{sum}}$ and $c^{G}_{\mathrm{sum}}$ are LLM-generated compressions of such subsequences. The distinction between the two types therefore reduces to \emph{which} part of $\mathcal{T}$ is drawn upon: local context is localized around $I_i$, either by position through $c^{L}_{i,\mathrm{surr}}$ or by reference through $\mathcal{R}_i$, whereas global context spans $\mathcal{T}$ as a whole. The next subsection describes how the two components of $\widetilde{C}_i$ are utilized across the stages of MMKG construction.

\subsection{Multi-Stage Context Utilization}
\label{subsec:utilization}
Having constructed $\widetilde{C}_i$, we next introduce how it can be used for MMKG construction. As illustrated in~\cref{fig:framework}, an MMKG is assembled along two branches that meet at modality fusion, and the constructed context can be injected at the two points marked by $\oplus$: when a visual element is turned into an image-based graph, and when that graph is fused with the text-based graph. Rather than incorporating the constructed context only once, we propose \textit{multi-stage context utilization}, where different construction stages utilize different components of $\widetilde{C}_i$ according to their specific objectives. For image-to-graph construction, both image-specific information and document-level context can facilitate a more comprehensive understanding of the visual element; therefore, we provide the full context $\widetilde{C}_i$ at this stage. In contrast, the fusion stage aims to establish correspondences between image- and text-derived entities. We therefore provide only the local context $c^{L}_i$, which provides more focused evidence for entity alignment.

\subsubsection{Context-Augmented Image-to-Graph Construction}
\label{subsec:i2g}
Images in documents are often difficult to interpret in isolation, as their semantics are closely related to both specific textual descriptions and the overall topic of the document. We therefore augment each visual element with both local and global contextual information during image-to-graph construction.

Existing methods build the image-based graph of a visual element $I_i$ as $\mathcal{G}^{\mathrm{v}}_i = f_{\mathrm{v}}(I_i, C_i)$, where $f_{\mathrm{v}}$ is an image-to-graph module, such as the Image2Graph module of MMGraphRAG~\citep{wan2025mmgraphrag}, and $C_i$ is the limited surrounding text discussed in ~\cref{sec:construction}. We leave the backbone $f_{\mathrm{v}}$ unchanged and replace only its context argument with the constructed context:
\begin{equation}
\mathcal{G}^{\mathrm{v}}_i
= f_{\mathrm{v}} \left( I_i,\; \widetilde{C}_i \right),
\label{eq:context_i2g}
\end{equation}
where $\mathcal{G}^{\mathrm{v}}_i$ denotes the resulting image-based entity-relation graph. This corresponds to the left $\oplus$ of~\cref{fig:framework}. Inside $f_{\mathrm{v}}$, the context conditions the MLLM that describes the visual element before entities and relations are extracted from that description, so a richer context propagates into every node and edge of $\mathcal{G}^{\mathrm{v}}_i$. Our modification is thus confined to the context supplied to the backbone.

The two levels of context provide complementary semantic information for visual knowledge extraction. Local context provides fine-grained evidence directly related to the target image, supporting the interpretation of image-specific concepts and relations. In contrast, global context provides a document-level semantic reference that connects the image to the broader topic of the document. By jointly incorporating both levels of context, the image-to-graph process can extract visual knowledge that is consistent with both image-specific textual evidence and the overall semantics of the document.

\subsubsection{Context-Guided Multimodal Knowledge Graph Fusion}
\label{subsec:fusion}
After constructing the image-based graphs, we explore how to better use the designed contexts to integrate them with the text-based knowledge graph extracted from the document. At this stage, the objective shifts from interpreting individual images to establishing correspondences between image- and text-derived knowledge. We utilize the local contexts as additional evidence to guide cross-modal knowledge integration.

Let $\mathcal{G}^{\mathrm{t}}$ denote the text-based knowledge graph constructed from document $\mathcal{D}$, and recall that $|\mathcal{I}| = N$. The unified MMKG is then obtained as
\begin{equation}
\mathcal{G_{M}}
= f_{\mathrm{fusion}} \left(
    \mathcal{G}^{\mathrm{t}},\;
    \{\mathcal{G}^{\mathrm{v}}_i\}_{i=1}^{N},\;
    \{c^{L}_i\}_{i=1}^{N}
  \right),
\label{eq:context_fusion}
\end{equation}
where $f_{\mathrm{fusion}}(\cdot)$ denotes the cross-modal knowledge integration process. Relative to the generic formulation in Eq.~\eqref{eq:generic_fusion}, the local contexts $\{c^{L}_i\}_{i=1}^{N}$ are supplied as an additional argument. This corresponds to the right $\oplus$ of~\cref{fig:framework}, and we now make explicit how $c^{L}_i$ acts there. Recall from ~\cref{sec:construction} that $f_{\mathrm{fusion}}$ aligns each visual entity with candidate textual entities derived from its associated textual context. In the backbone method, this context is limited to text chunks surrounding $I_i$, which may exclude relevant information distributed elsewhere in the document. We therefore augment the available context with the designed local context $c_i^L$, which incorporates relevant information beyond these surrounding chunks. This allows a broader range of relevant textual entities to be considered as potential alignment candidates. The LLM then determines the final alignment following the original fusion procedure. In this way, $c_i^L$ facilitates the alignment of image-derived entities with their counterparts in the text-based knowledge graph, thereby supporting more effective cross-modal knowledge integration and reducing ambiguity during modality fusion.

Across the two stages, the constructed context serves two complementary roles: it supports a more comprehensive understanding of each visual element during Image2Graph construction and provides relevant textual evidence for aligning the resulting visual entities with their textual counterparts during modality fusion.

\begin{table*}[t]
    \centering
    \caption{Results of different textual context configurations on the VisionHeavy subset of MMLongBench-Doc. Each cell reports strict accuracy / soft accuracy (\%). Figure, Table, and Chart labels may overlap; Misc contains questions without any of these labels. Bold and underlined values indicate the best and second-best results, respectively.
    }
    \label{tab:overall}
    \small
    \setlength{\tabcolsep}{3pt}
    \begin{tabular}{lllccccc}
        \toprule
        Method & Extra Local Context & Global Context
        & Figure & Table & Chart & Misc & Overall \\
        \midrule

                 Direct Inference
        & --
        & --
        & 6.25 / 6.25
        & 4.00 / 4.00
        & 0.00 / 0.00
        & 21.95 / 21.95
        & 11.32 / 11.32 \\
        
        MMGraphRAG
        & --
        & --
        & 6.25 / 6.25
        & \underline{20.00} / 22.46
        & \textbf{31.25} / \textbf{31.25}
        & 34.15 / 34.15
        & 23.58 / 24.17 \\

        \midrule

        \multirow{3}{*}{CEMMKG}
        & Ref. paragraph
        & Doc-level
        & 12.50 / 12.50
        & 16.00 / 16.00
        & \textbf{31.25} / \textbf{31.25}
        & 26.83 / 26.83
        & 22.64 / 22.64 \\

        & Ref. paragraph summary
        & Doc-level
        & \underline{15.63} / \underline{15.63}
        & \textbf{24.00} / \textbf{26.46}
        & \underline{25.00} / \underline{25.00}
        & \underline{43.90} / \underline{46.76}
        & \underline{31.13} / \underline{32.82} \\

        & Ref. sentence
        & Doc-level
        & \textbf{21.88} / \textbf{24.70}
        & \textbf{24.00} / \underline{24.00}
        & \textbf{31.25} / \textbf{31.25}
        & \textbf{51.22} / \textbf{54.01}
        & \textbf{34.91} / \textbf{36.84} \\

        \bottomrule
    \end{tabular}
\end{table*}

\section{Experiments}
In this section, we evaluate CEMMKG through comprehensive experiments to address the following research questions: (i) (\textbf{RQ1}): How do different textual context designs affect the performance of MMKG-based RAG? (ii) (\textbf{RQ2}): Can CEMMKG be effectively applied to different MMKG-based RAG methods? (iii) (\textbf{RQ3}): How do the designed context types and context utilization stages affect the performance of CEMMKG?

\subsection{Experimental Setup}
\subsubsection{Datasets}

We evaluate our method on MMLongBench-Doc~\citep{ma2024mmlongbenchdoc}, DocBench~\citep{zou2025docbench}, and SPIQA~\citep{pramanick2024spiqa}. We use the test-A of SPIQA. For DocBench, we use papers and questions in the Academia domain. For MMLongBench-Doc, a portion of the questions are either unanswerable or can be answered without visual information. Therefore, we select a subset named \emph{MMLongBench-Doc-VisionHeavy} from MMLongBench-Doc. More details can be found in~\cref{suppsec:_datasert_details} of the supplementary material.

\subsubsection{Baseline Methods}
We include the following two types of methods as baselines for comparison. \textbf{Direct Inference}: A multimodal large language model directly answers the question based on the provided document. \textbf{MMKG-based RAG}: Multimodal RAG methods that can construct unified multimodal knowledge graphs and perform retrieval over the resulting graph. We consider RAG-Anything~\citep{guo2025raganything} and MMGraphRAG~\citep{wan2025mmgraphrag} in our work.

\subsubsection{Evaluation Metrics}
For the VisionHeavy subset of MMLongBench-Doc, we follow the official evaluation protocol~\citep{ma2024mmlongbenchdoc}. An LLM first extracts a short-form answer from each response, which is then evaluated using the benchmark's rule-based scorer. We refer to the mean per-question score as \emph{soft accuracy} and additionally report \emph{strict accuracy}, defined as the proportion of questions receiving full credit (a score of 1). For SPIQA and DocBench Academia, we use an LLM to evaluate each generated answer against its reference answer. The judge model assigns a binary correctness label to each response, and we report accuracy. All metrics are reported as percentages.

\subsubsection{Implementation Details}
All experiments are conducted on servers equipped with four NVIDIA A6000 48GB GPUs. Following previous works~\citep{wan2025mmgraphrag,guo2025raganything}, we use MinerU~\citep{wang2024mineru} to parse PDF documents and extract content from different modalities. In our experiments, we adopt both RAG-Anything~\citep{guo2025raganything} and MMGraphRAG~\citep{wan2025mmgraphrag} as backbone frameworks to demonstrate the effectiveness of the proposed CEMMKG. To ensure a fair comparison, we use the same retrieval procedure and hyperparameters across all CEMMKG context configurations for each backbone framework. For QA task, the judge LLM is Llama-3.1-70B-Instruct. More details are in~\cref{suppsec:more_implementation_details} of the supplementary material. 

\begin{table}[t]
  \centering
  \caption{Overall accuracy (\%) on SPIQA (test-A) and DocBench Academia, respectively.}
  \label{tab:overall_more_bench}
  \small
  \setlength{\tabcolsep}{4pt}
  \begin{tabular}{@{}lccc@{}}
    \toprule
    Dataset & MMGraphRAG & CEMMKG & $\Delta$ \\
    \midrule
    SPIQA (test-A)    & 60.25 & \textbf{66.82} & +6.57 \\
    DocBench Academia & 60.67 & \textbf{71.13} & +10.46 \\
    \bottomrule
  \end{tabular}
\end{table}

\subsection{Multimodal GraphRAG Performance}
\subsubsection{Overall Performance}
\label{sec:overall_results}
To answer \textbf{RQ1}, we conduct comprehensive experiments based on different textual context configurations.~\cref{tab:overall} presents the overall performance of using different textual context designs on the VisionHeavy subset. Among all evaluated configurations, incorporating the reference sentence(s) as additional local context together with document-level global context delivers the best overall performance, achieving a strict accuracy of 34.91\% and a soft accuracy of 36.84\%. This substantially outperforms MMGraphRAG, which achieves 23.58\% strict accuracy and 24.17\% soft accuracy. Similarly, when we use the reference paragraph summary, the performance is also strong with a strict accuracy of 31.13\% and a soft accuracy of 32.82\%. These results demonstrate that incorporating appropriately designed textual context during MMKG construction can effectively improve downstream multimodal GraphRAG performance.

However, we can observe that directly using the full reference paragraph(s) as additional local context achieves an overall soft accuracy of only 22.64\%, even lower than the 24.17\% of MMGraphRAG. This result suggests that simply introducing more textual context does not necessarily lead to better performance. The possible reason is that the complete paragraph(s) could contain information that is only weakly related to the corresponding visual content, introducing irrelevant or redundant context during MMKG construction. This observation indicates that the effectiveness of local context depends not only on the amount of contextual information provided, but also on its granularity and information density.

Overall, these results underscore the importance of choosing an appropriate level of textual granularity for high-quality MMKG construction and improved downstream multimodal GraphRAG performance. The results in~\cref{tab:overall_more_bench} further show the effectiveness of our proposed CEMMKG on more datasets.

\subsubsection{Performance across Different Types}

To better understand the effectiveness of CEMMKG, we further conduct experiments according to the types of visual content involved in each question, including Figure, Chart, and Table. Please note that these labels are not mutually exclusive, as a single question may involve multiple types of visual content. The per-type results are reported to provide a more fine-grained understanding of CEMMKG performance. As shown in~\cref{tab:overall}, the reference-sentence context brings the most substantial improvement on Figure-related questions, increasing soft accuracy from 6.25\% to 24.70\%. For Table-related questions, the reference paragraph summary achieves the best soft accuracy of 26.46\%, compared with 22.46\% for MMGraphRAG. These results suggest that different textual context designs could provide different benefits depending on the category of visual information involved.

Interestingly, the proposed context design also yields substantial improvements on the Misc questions, increasing soft accuracy from 34.15\% to 54.01\% and strict accuracy from 34.15\% to 51.22\%. Misc refers to questions without Figure, Chart, or Table labels, but these questions may still involve multimodal information. One possible explanation for this improvement is that the designed contextual information enhances not only the representation of individual image-based graphs but also the modality integration. These improvements may therefore also benefit questions that rely on textual knowledge or cross-modal semantic connections, even when they do not explicitly require visual content.

\begin{table}[t]
    \centering
    \caption{Performance of CEMMKG with RAG-Anything as the backbone under different additional local context configurations. Results are reported as strict and soft accuracy (\%).}
    \label{tab:generalizability}
    \small
    \setlength{\tabcolsep}{3pt}
    \begin{tabular}{@{}llcc@{}}
        \toprule
        Method & Extra local context & Strict & Soft \\
        \midrule
        RAG-Anything & -- & 28.89 & \underline{36.18} \\
        \midrule
        CEMMKG & Ref. paragraph
            & 28.89 & 34.49 \\
        CEMMKG & Ref. paragraph summary
            & \underline{31.11} & 33.13 \\
        CEMMKG & Ref. sentence
            & \textbf{35.56} & \textbf{41.12} \\
        \bottomrule
    \end{tabular}
\end{table}

\subsection{Applicability to Different Methods}
\cref{tab:overall} reports the performance of CEMMKG with MMGraphRAG as the backbone. To investigate whether CEMMKG is also applicable to other multimodal GraphRAG methods (\textbf{RQ2}), we further conduct experiments with RAG-Anything. We integrate CEMMKG into RAG-Anything's visual representation stage by augmenting the description prompt for each image or table with the document-level summary and any available explicit reference sentence(s) before generating the structured visual description. The resulting description is then processed by RAG-Anything's original multimodal graph construction pipeline. Unlike MMGraphRAG, RAG-Anything already supplies each visual with native contextual text that can span neighboring pages. Therefore, to better evaluate the effectiveness of the designed textual context, we evaluate CEMMKG on a subset of VisionHeavy questions that require cross-page references. As shown in~\cref{tab:generalizability}, the original RAG-Anything achieves a strict accuracy of 28.89\% and a soft accuracy of 36.18\%. Incorporating our designed reference sentence(s) as extra contextual information further improves these results to 35.56\% and 41.12\%, respectively, demonstrating its effectiveness even on a strong MMKG-based RAG baseline. We also observe a trend consistent with that seen when using MMGraphRAG as the backbone: using reference sentences as local context achieve the best performance.

To evaluate the constructed MMKG quality beyond downstream accuracy, we evaluate the alignments produced by fusion directly. The results are in~\cref{sec:supp_align}.

\subsection{Ablation Studies}
To address \textbf{RQ3}, we conduct two ablation studies to examine the effects of context types and local context utilization stages on CEMMKG's performance.

\paragraph{Effect of Context Types.}
As shown in ~\cref{tab:supp_component}, adding either component improves performance over MMGraphRAG, while combining both yields the highest strict and soft accuracy. These results suggest that the two context types provide complementary benefits.

\begin{table}[t]
    \centering
    \caption{Context Type ablation on VisionHeavy with MMGraphRAG as the backbone. Ref. and Global denote additional local context and document-level global context, respectively. Results are strict and soft accuracy (\%).}
    \label{tab:supp_component}
    \small
    \setlength{\tabcolsep}{4pt}
    \begin{tabular}{@{}lcccc@{}}
        \toprule
        Variant & Ref. & Global & Strict & Soft \\
        \midrule
        MMGraphRAG
        & $\times$ & $\times$
        & 23.58 & 24.17 \\

        + Local Context
        & $\checkmark$ & $\times$
        & 26.42 & 27.53 \\

        + Global Context
        & $\times$ & $\checkmark$
        & 28.30 & 30.42 \\

        CEMMKG
        & $\checkmark$ & $\checkmark$
        & \textbf{34.91} & \textbf{36.84} \\
        \bottomrule
    \end{tabular}
\end{table}

\begin{table}[t]
    \centering
    \caption{Stage ablation of local context on VisionHeavy. Global context is used only in Image2Graph stage. Results are strict and soft accuracy (\%).}
    \label{tab:ablation_stage}
    \small
    \setlength{\tabcolsep}{3pt}
    \begin{tabular}{@{}lcccc@{}}
        \toprule
        Variant & Image2Graph & Fusion & Strict & Soft \\
        \midrule
        w/o Local Context.\
        & $\times$ & $\times$
        & 28.30 & 30.42 \\

        Image2Graph only
        & $\checkmark$ & $\times$
        & 31.13 & 32.52 \\

        CEMMKG
        & $\checkmark$ & $\checkmark$
        & \textbf{34.91} & \textbf{36.84} \\
        \bottomrule
    \end{tabular}
\end{table}

\paragraph{Effect of Context Utilization Stages.} CEMMKG utilizes the designed local context during both Image2Graph and modality fusion. As shown in~\cref{tab:ablation_stage}, using local context in Image2Graph improves both strict and soft accuracy. Extending its use to modality fusion stage yields further gains. These results demonstrate the effectiveness of incorporating local context at both stages.

\section{Conclusion}
In this work, we systematically investigate the role of textual context in GraphRAG-oriented MMKG construction. We introduce CEMMKG, a framework that systematically defines and utilizes textual context in both visual information processing and modality fusion. Specifically, CEMMKG designs textual context for visual elements from complementary perspectives, including local context at different granularities and global context based on different document-level representations. We further explore how the designed context can be effectively incorporated across different stages of MMKG construction. Extensive experiments based on different context configurations demonstrate that appropriately designed textual context can effectively improve the performance of MMKG-based RAG. Moreover, CEMMKG can be effectively applied to different MMKG-based RAG methods, demonstrating its broader applicability. Overall, our findings highlight the importance of explicitly considering and leveraging contextual information when constructing MMKGs for multimodal Graph RAG. Future work could extend our framework to a broader range of modalities, such as video and audio, and investigate how contextual information should be defined and utilized for different modality combinations.
\section*{LLM Usage Statement}
The authors used ChatGPT for writing assistance, and all AI-generated content was carefully reviewed and verified by the authors. ChatGPT was also used to generate a few icons used in some of the figures for visualization purposes. Additionally, Codex and Claude Code were used to assist with parts of the implementation and code review.

{
    \small
    \bibliographystyle{ieeenat_fullname}
    \bibliography{main}
}

\clearpage
\maketitlesupplementary

\section{Dataset Details}
\begin{table}[t]
  \centering
  \caption{Statistics of datasets after the exclusion rules described in Section~\ref{suppsec:_datasert_details}.}
  \label{tab:supp_bench_data}
  \small
  \setlength{\tabcolsep}{3pt}
  \begin{tabular}{lcc}
    \midrule
    Benchmark & Papers & Questions\\
    \midrule
    SPIQA (test-A)    & 114 & 639 \\
    DocBench Academia  &  47 & 239 \\
    \bottomrule
  \end{tabular}
\end{table}

\label{suppsec:_datasert_details}
The test-A split of SPIQA consists of 118 papers and 666 questions. We removed one duplicate question, resulting in 665 questions. DocBench Academia has 49 papers and 303 questions. We exclude the meta-data category in DocBench, because it asks about properties of a document rather than its content. We also exclude documents for which MMKG-based RAG methods fail to construct an MMKG, as well as questions for which they fail to generate an answer. The statistics are shown in Table~\ref{tab:supp_bench_data}.

The selection of MMLongBench-Doc-VisionHeavy follows two criteria: First, a document should be \emph{vision-intensive}, meaning that a high proportion of its answerable questions require evidence from visual information such as figures. Second, a document should contain \emph{non-local evidence dependencies}, meaning that answering its questions requires connecting visual elements with evidence beyond their neighboring textual context. The resulting VisionHeavy subset contains 106 questions. Among the answerable questions, 80.2\% require visual information, and 43.2\% rely on information distributed across multiple pages. The selected subset spans six of the seven document types in the original MMLongBench-Doc benchmark~\citep{ma2024mmlongbenchdoc}, including academic paper, administration\&industry files, brochure, guideline, research report, and tutorial/workshop.

\section{Additional Implementation Details}
\label{suppsec:more_implementation_details}
When using MMGraphRAG as the backbone, we employ Qwen2.5-72B-Instruct-AWQ~\citep{qwen2025qwen25report} for text-based knowledge graph construction, text summarization, and textual generation. InternVL2.5-38B-MPO-AWQ~\citep{chen2024internvl25} is used to generate image descriptions, and perform multimodal question answering. We use deterministic decoding across all generation stages, setting the temperature to 0 and top-p to 1. We set the maximum output length of the answer extractor to 1,024 tokens. We retain up to five explicit textual references for image description and up to three for graph fusion. 

\section{Quality of Cross-Modal Alignment}
\label{sec:supp_align}
To directly evaluate cross-modal alignment quality, we randomly sample 100 aligned image–text entity pairs from the MMKGs constructed by MMGraphRAG and CEMMKG across 14 VisionHeavy documents.

An LLM-based judge model evaluates each alignment using the corresponding image, caption, and entity pair, classifying it as \emph{correct} (referring to the same entity or concept), \emph{related} (associated but distinct), or \emph{incorrect} (unrelated or unsupported by the image). We define alignment precision as the proportion of pairs classified as \emph{correct}. CEMMKG achieves an alignment precision of 49\%, compared with 42\%  for MMGraphRAG. These results suggest that our proposed CEMMKG not only improves downstream QA performance but also achieves higher cross-modal alignment precision. The judge model for cross-modal alignment is OpenAI Codex.

\end{document}